\documentclass[10pt]{article}
\usepackage[utf8]{inputenc}
\usepackage{xspace}
\usepackage[a4paper, total={5in, 7in}]{geometry}
\usepackage{paralist}
\usepackage{tabularx}
\usepackage{booktabs}
\usepackage{graphicx}
\usepackage{amsthm,amssymb}
\usepackage{natbib}
\usepackage[dvipsnames]{xcolor}
\usepackage{enumitem}
\usepackage{algorithm}
\usepackage{authblk}
\usepackage{algpseudocode}  
\usepackage{amsmath} 
\usepackage{algorithmicx}
\usepackage{hyperref}
\usepackage{color,soul}

\usepackage{todonotes}

\title{MICROTRIPS: MICRO-geography TRavel Intelligence and Pattern Synthesis}

\author{Yangyang Wang}
\author{Tayo Fabusuyi\thanks{\textit{Corresponding author}, Fabusuyi@umich.edu}}
\affil{{University of Michigan, Ann Arbor}}

\date{}

\begin{document}

\maketitle
\section*{Abstract}

This study presents a novel small-area estimation framework to enhance urban transportation planning through detailed characterization of travel behavior. Our approach improves on the four-step travel model by employing publicly available microdata files and machine learning methods to predict travel behavior for a representative, synthetic population at small geographic areas. This approach enables high-resolution estimation of trip generation, trip distribution, mode choice, and route assignment. Validation using ACS/PUMS work-commute datasets demonstrates that our framework achieves higher accuracy compared to conventional approaches. The resulting granular insights enable the tailoring of interventions to address localized situations and support a range of policy applications and targeted interventions, including the optimal placement of micro-fulfillment centers, effective curb-space management, and the design of more inclusive transportation solutions particularly for vulnerable communities.

\hfill\break%
\noindent\textit{Keywords}: Small Area Estimation; Four Step Travel Model (FSTM); Synthetic Populations; Inclusive Transportation; Policy Optimization
\newpage

\bibliographystyle{plainnat}
\section{Introduction and Background Context}
This study seeks to advance the field of travel behavior analysis by strengthening its foundational frameworks and embracing innovative data integration techniques. Leveraging the well - established Four Step Travel Model (FSTM) \cite{McNally2000Four} — a widely recognized approach encompassing trip generation, trip distribution, mode choice, and route assignment—we introduce a robust methodology that both captures the nuanced complexities of urban travel and enhances predictive accuracy. Our approach goes beyond traditional models by demonstrating how various factors shape travel patterns, particularly within specific segments of the population and at a fine-grained geographic scale.

We have enriched the traditional FSTM by incorporating more robust modeling methods that are particularly well-suited for small-area estimations. A key element of our approach are Random Utility (RU) \cite{McFadden1973Conditional} \& \cite {McFadden1981Econometric} and Discrete Choice Models (DCMs) \cite{BenAkiva1985Discrete} \& \cite{Train2009Discrete}, which are crucial for capturing individual preferences and choices. These models allow us to analyze the decision-making processes underlying travel behavior, offering a detailed understanding of how factors such as cost, time, and convenience influence mode and route selection. By integrating these established models with advanced data techniques, our approach significantly improves predictive accuracy and policy relevance, demonstrating clear value over the conventional FSTM framework.

Recognizing that the quality of our findings depends on the data used, we carefully selected the datasets for this research. Our study draws on regional household travel surveys, such as the Puget Sound Regional Commission (PSRC) Household Travel Survey, as well as publicly available microdata like the U.S. Census Public Use Microdata Sample (PUMS), which underlies the American Community Survey (ACS) tables. Together, these datasets provide a solid foundation for our analysis by offering comprehensive insights into demographic, socioeconomic, and travel behavior characteristics. Leveraging this rich data, we employ a combination of raking and machine learning methods to develop a replicable and reproducible approach for generating high-resolution insights on individuals’ travel behavior – demand, distribution, routes and mode - at fine geographic scale. By integrating traditional statistical techniques with advanced computational methods, our approach guarantees that demand estimates are both accurate and reliable.

A key innovation in our approach is the development of a framework that leverages synthetic population generated through an iterative proportional fitting (IPF) method, ensuring that our estimates reflect the diverse and dynamic nature of populations of interest. By offering a richer assessment of the local travel pattern for specific population cohorts, the study highlights the primary drivers of change, the directions and magnitudes of these changes, the factors that may be amenable to policy modification and how sensitive these factors are relative to each segment of the population. Our approach addresses critical gaps in localized data necessary for informed decision-making, particularly at the neighborhood or sub-city level where many interventions are typically implemented. The insights provided by this study are invaluable for municipalities seeking to understand and manage the rapidly evolving patterns and trends in travel behavior. A crucial benefit of this approach is that it is not data intensive, and it could easily be tailored to address niche populations.

We showcase our approach using a demo webpage \url{https://seattle-travel.replit.app/} that provides an overview of how small area estimates are generated using Seattle as a case study.  Though the demonstration site is on transportation, the approach could easily be extended to multiple domain areas. The webpage presents estimated travel mode choices, destinations, trip volume and routes in a given period at the Census Block Group (CBG) level for Seattle, thus providing crucial insights that could be localized to specific geographical areas, and for designing targeted policy responses. Our research offers more reliable and actionable insights and provides results at a level that makes it practical for city planners, municipal governments, and community-based organizations to track and respond to local trends effectively, ultimately fostering better-targeted interventions.


\section{Literature Review}
\subsection*{Existing Works}
Recent advances in transportation research highlight the complex interplay between the built environment, travel patterns, traffic dynamics, zoning ordinances, and land use policies. This progress is characterized by the integration of diverse methodologies and a growing emphasis on data-driven approaches, which provide increasingly detailed and nuanced insights for small geographic areas.

A rich body of literature has consistently documented the profound influence of urban form and individual travel patterns on overall mobility. \cite{Stead2001Influence} provided a foundational understanding of how different urban layouts influence travel behaviors, including mode choice and trip frequency. This viewpoint is further complemented by studies such as \cite{Goulet-Langlois2017Measuring}, which measures the regularity of individual travel patterns using intelligent transportation system (ITS) data, and \cite{Peirce2003Impact}, that explores how real-time traveler information impacts travel decisions . These works collectively emphasize the importance of both static urban design factors and dynamic information flows in shaping travel choices, establishing the groundwork for more adaptive planning models.

The application of machine learning has significantly transformed the field of transportation research, offering sophisticated tools for analyzing and predicting travel behaviors and traffic patterns. \cite{Gong2018Trip} and \cite{Hagenauer2017Machine} exemplify this by showcasing the use of machine learning to identify trip purposes and model travel mode choices, respectively, thereby enhancing the accuracy of travel behavior analysis . Machine learning's capacity for handling non-linear relationships and complex interactions has proven invaluable for capturing the nuanced factors influencing human travel behavior, moving beyond the limitations of traditional linear models. 

Concurrently, network theory and advanced mathematical frameworks have been increasingly applied to traffic analysis. \cite{Solé-Ribalta2016Congestion} utilized network theory to identify congestion hotspots, while \cite{Hu2022Traffic} and \cite{Zhang2022Complex} leveraged complex network information for traffic flow prediction. This interdisciplinary approach integrates sophisticated mathematical frameworks, such as Hodge theory in \cite{Jiang2011Topological}, \cite{Lim2020Hodge}, and \cite{Aoki2022Dynamic}, providing a comprehensive understanding of urban spatial structures and traffic dynamics . These theoretical advancements offer deeper insights into the underlying mechanisms of urban mobility, providing a more rigorous foundation for developing innovative solutions in transportation planning and management.

\subsection*{Identified Gaps}

Despite significant advancements in transportation research, a synthesis of existing works reveals several critical gaps that limit the comprehensiveness and actionable utility of current models. Addressing these limitations is paramount for developing truly effective and responsive urban planning strategies.

One notable gap lies in the insufficient integration of static urban design factors with dynamic real-time data. While the impact of urban form on travel patterns is well-established, there remains a need to seamlessly integrate these static elements with real-time data acquisition, as explored by \cite{Stead2001Influence} and \cite{Peirce2003Impact}. Future research is necessary to create adaptive urban planning models that effectively combine these influences, enhancing the responsiveness of urban infrastructure to changing travel behaviors.   

A second limitation is the lack of comprehensive models for individual travel patterns. Although progress has been made in understanding individual travel regularity by \cite{Goulet-Langlois2017Measuring} and trip purposes by \cite{Gong2018Trip}, a void exists in developing truly comprehensive models that integrate individuals' travel behavior with broader urban influences and real-time data. Such models are crucial for capturing the full spectrum of factors influencing individual travel decisions.

A third critical gap relates to the translation of advanced predictive models into prescriptive solutions. Advanced machine learning and network theory applications from \cite{Hagenauer2017Machine}, \cite{Hu2022Traffic}, and \cite{Zhang2022Complex} have primarily focused on predictive modeling and traffic analysis. However, further development is needed to move beyond mere prediction to prescribing solutions for optimizing travel patterns and reducing congestion through proactive urban planning and policy-making.  

These shortcomings are what informed the present study. What we are proposing introduces a groundbreaking small-area estimation framework that bridges critical gaps in urban transportation planning by integrating urban design with dynamic real-time data and developing comprehensive individual-level models. It uses publicly available microdata and machine learning to predict travel behavior for synthetic populations at fine geographic scales such as census block groups, addressing the "resolution gap" of traditional, coarse-grained models. While not explicitly focused on environmental outcomes, the framework implicitly contributes to understanding environmental and behavioral linkages by providing granular data that can inform tailored interventions and be fed into emissions estimation tools. Furthermore, the research moves beyond mere prediction to offer prescriptive solutions, enabling city planners to analyze scenarios, anticipate impacts, and implement targeted interventions for optimized travel patterns and reduced congestion. Crucially, the study prioritizes ethical considerations by employing synthetic populations, which provide the benefits of highly detailed individual-level data for accurate modeling while safeguarding personal privacy.

\section{Empirical Framework}

The project's empirical framework, while borrowing heavily from the the classical Four-Step structure, demonstrates a significant evolution beyond its traditional, highway-centric origins. The methodologies and data sources employed, such as OpenStreetMap (OSM) for network data, General Transit Feed Specification (GTFS) for transit information, and the Puget Sound Regional Council's (PSRC) highlight a sophisticated adaptation. This shows that our approach, illustrated using Seattle as the testbed, is not a rigid application of the 1950s paradigm but a contemporary, advanced iteration that incorporates multimodal travel and a more nuanced understanding of traveler behavior, moving towards activity-based principles within the established framework. This procedural approach significantly enhances its relevance for addressing complex modern urban planning challenges, including traffic congestion, environmental sustainability, and equitable access to transportation.

\subsection{Data and Data Sources}
\begin{figure}
\begin{center}
\includegraphics[width=0.8\textwidth]{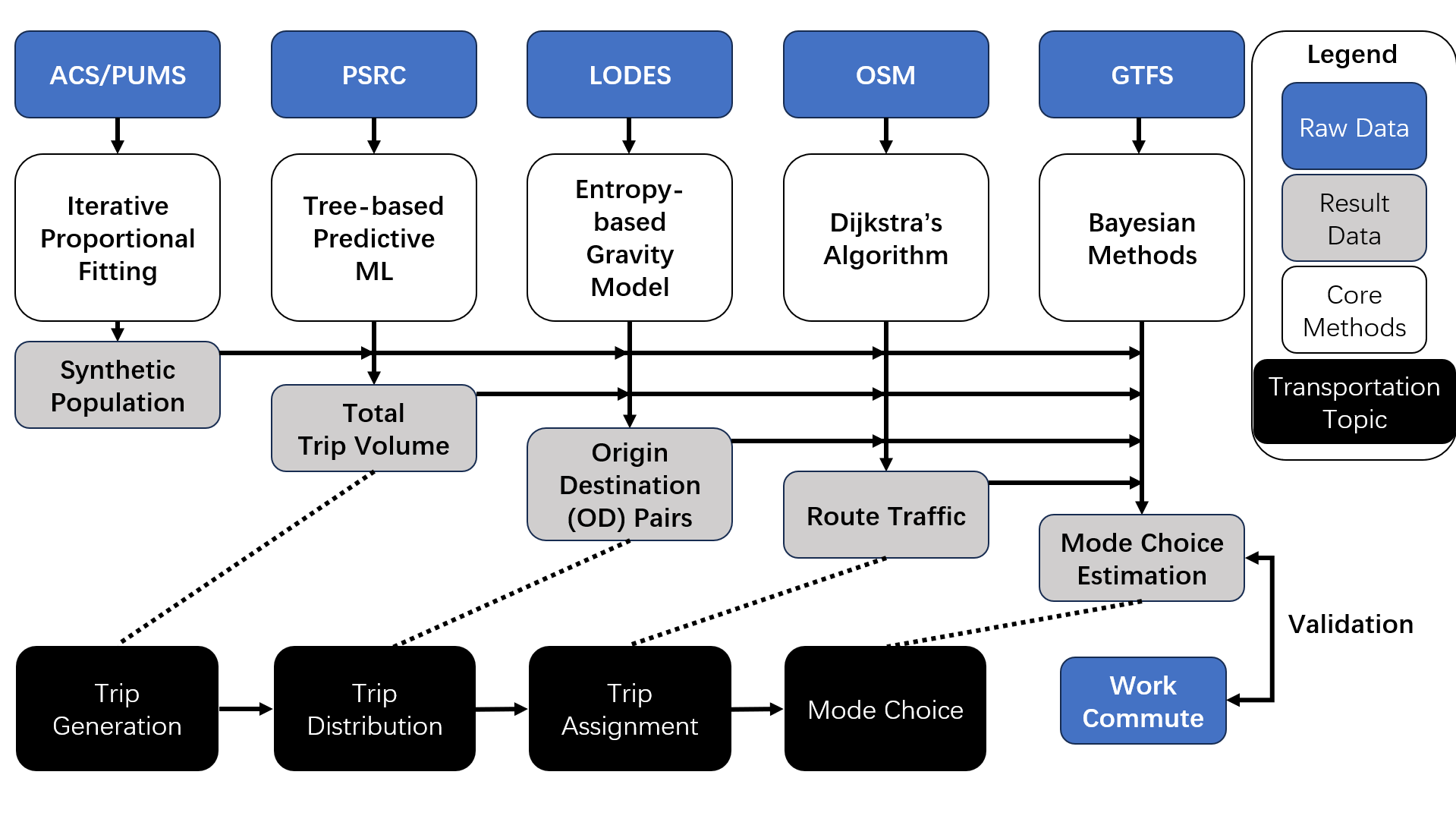}
\caption{Panoramic Chart with Four Step Travel Demand Model}
\label{fig:panoramic_chart}
\end{center}
\end{figure}

The FSTM, a framework that was developed in the 1950s primarily for for highway planning \citep{McNally2000Four} represents the seminal empirical model for transportation forecasting. Over the decades, this foundational model has undergone significant evolution, expanding its scope in the 1970s and 1980s to encompass multimodal trips and integrate more sophisticated modeling techniques. This continuous adaptation underscores its enduring role as a ubiquitous tool for establishing a critical link between land use patterns and travel behavior, thereby informing comprehensive transportation planning efforts.

\begin{table}[htbp]
\centering
\renewcommand{\arraystretch}{1.4}
\begin{tabular}{|p{4cm}|p{2.5cm}|p{5cm}|}
\hline
\textbf{Transportation Dimension} & \textbf{Core Source Data} & \textbf{Core Methodology} \\
\hline

Trip Generation & PSRC & Tree-based Predictive ML \\
\hline

OD Distribution & LODES & Entropy-based Gravity Model \\
\hline

Route Assignment & OSM & Dijkstra's Algorithm \\
\hline

Mode Choice & GTFS & Bayesian Methods \\
\hline

Syntetic Population & ACS/PUMS & Iterative Proportional Fitting \\
\hline

\end{tabular}
\end{table}

Table 1 provides a high-level description of the transportation dimensions being analyzed and for each, the core methodology utilized and the primary data used. More information on the datasets is provided below.

\begin{table}[h!]
\centering
\begin{tabular}{|p{2.5cm}|p{3.5cm}|p{5.5cm}|}
\hline
\textbf{Data Source} & \textbf{URL / Citation} & \textbf{Brief Description} \\
\hline
PSRC Data & \cite{PSRCDatasource} & Provides household, demographic, and trip survey data used in trip generation and mode choice modeling. Captures income, household size, vehicle availability, and neighborhood characteristics. \\
\hline
LODES Data & \cite{LODESData} & U.S. Census Bureau dataset detailing home-to-work commuting flows at fine spatial scales (OD, RAC, WAC). Supports accurate modeling of employment-driven trips. \\
\hline
OpenStreetMap (OSM) & \cite{OpenStreetMap} & Open-source geospatial data for building the multimodal transport network. Used to construct route graphs and shortest-path calculations across modes. \\
\hline
GTFS Data & \cite{GTFSData} & Standardized transit feed including schedules, routes, and stops. Enables spatiotemporal transit network modeling and supports mode choice analysis. \\
\hline
ACS / PUMS & \cite{ACSPUMS} & Provides demographic and household microdata for population synthesis. Used to generate synthetic populations reflecting real-world demographics. \\
\hline
Work Commute Data & \cite{WorkCommuteData} & ACS commuting statistics used for validation. Includes mode of travel, departure times, travel times, and residence–workplace locations. \\
\hline
\end{tabular}
\caption{Summary of primary datasets used in the study.}
\end{table}

\subsection{Modeling Workflow}
\paragraph{Synthetic Population Generation}
Population synthesis is a fundamental process that involves generating a synthetic population by expanding disaggregate sample data, such as Public Use Microdata Sample (PUMS) or travel surveys, to accurately reflect known aggregate distributions derived from census summary files. This step is crucial as it provides the necessary input for activity-based models (ABMs), which simulate the travel choices of individual households and persons rather than merely aggregate trips.   

The methodology for synthetic population generation typically employs Iterative Proportional Fitting (IPF) to create joint count data from aggregate Census/ACS survey data (\cite{DemingStephan1940}). Subsequently, representative households and persons are sampled directly from PUMS files. PUMS files contain disaggregate records for persons and households, which can be linked to provide a comprehensive microdata foundation. 

The process of synthetic population generation bridges aggregate data to disaggregate behavior. While Census data provides aggregate distributions of population characteristics, PUMS offers the underlying microdata for individual households and persons. Synthetic population generation, particularly through techniques like Iterative Proportional Fitting (IPF), acts as the critical link, allowing the creation of a full, disaggregate population that statistically matches the known aggregate controls. This enables the fundamental shift from modeling abstract "trips" to simulating the nuanced "travel choices of households and persons". This capability allows the Seattle model to capture behavioral heterogeneity and individual-level nuances that aggregate models inherently miss. Instead of assuming average travel behavior for an entire census tract, the model can simulate how distinct household types (e.g., single-person, multi-generational families, varying income brackets) and individual characteristics (e.g., age, worker status, vehicle availability) make different travel decisions. This leads to a richer, more realistic simulation of urban mobility, which is essential for predicting "travel trails for each type of individual" as specified in the user query.  

\paragraph{Machine Learning for Trip Count Prediction}
Trip generation constitutes the initial phase in the conventional four-step transportation forecasting, primarily focusing on predicting the number of trips each type of socio-demographic individual would make.

The application of machine learning for trip count prediction leverages ML's capacity for non-linearity and behavioral complexity. Traditional trip generation models, such as linear regression or cross-classification, inherently assume linear relationships or fixed categorical influences. However, human travel behavior is a highly complex and non-linear phenomenon, influenced by a multitude of interacting socioeconomic, demographic, and environmental factors. Machine learning models, particularly ensemble tree-based methods (Random Forests, Gradient Boosting) and deep learning architectures, are uniquely adept at identifying and capturing these intricate, non-linear relationships and complex interactions without requiring explicit prior assumptions about their functional forms (\cite{Breiman2001RandomForests}). By leveraging ML for trip count prediction, the Seattle model can achieve significantly higher accuracy and greater behavioral realism in its foundational trip estimates.

\paragraph{Network Model for Census Block Groups and Travel Trail Prediction}

The integration of predicted trip counts proceeds as follows: the output from the trip generation phase (predicted trip counts or trip ends) serves as the primary input for the trip distribution step, which then matches trip origins with their respective destinations. This process culminates in the creation of origin-destination (O-D) matrices, typically categorized by trip purpose. In this study, an entropy-based gravity model is employed for O-D pairing (\cite{Wilson1970Entropy}). This model, rooted in the principle of maximizing the uncertainty of the trip distribution subject to known constraints (total productions, total attractions, and total travel cost), provides a statistically sound method for estimating the most probable distribution of trips between all possible origin-destination pairs. This macro-level approach effectively connects the "predicted count of trips from the ML model" (derived from the trip generation phase) with the network characteristics, generating a comprehensive matrix of travel demand between all TAZs. These O-D demands are subsequently assigned to the traffic network in the "traffic assignment" step, which determines the routes travelers choose.

The "Predicted Travelling Trail for Each Type of Individual" in this context refers to the outcome of the trip assignment (or route assignment) phase, where the previously determined Origin-Destination (O-D) demands are allocated onto the specific links of the network. For each O-D pair, Dijkstra's shortest path algorithm is employed to determine the optimal "travel trail" or route within the urban traffic network. Dijkstra's algorithm efficiently identifies the path with the minimum accumulated impedance (e.g., travel time, distance, cost) between two nodes in a graph (\cite{Dijkstra1959Note}). This step is crucial for understanding how travelers actually utilize the infrastructure.


\paragraph{Bayesian Methods for Travel Method Analysis and Refinement}

Mode choice is a complex decision influenced by a wide array of factors, including traveler characteristics (e.g., age, gender, income, driver's license status, vehicle availability), modal availability and characteristics (e.g., in-vehicle time, walk/access time, initial wait time, number of transfers, fare, parking cost), journey characteristics (e.g., time of day, presence of stops on a tour), and land use characteristics (e.g., population density, employment density, measures of mixed land use density, intersection density).

The application of Bayesian methods to determine probabilities of different transportation methods signifies a move beyond simple deterministic predictions (e.g., "X trips will use transit") to a more nuanced, probabilistic understanding of mode choice. Bayesian methods are inherently suited for this, as they naturally integrate and quantify uncertainty in their predictions. This contrasts sharply with many traditional models that provide a single "best" estimate without indicating the confidence level. For transportation planners in Seattle, this translates into a risk-informed decision-making framework. Instead of relying on a single predicted value, they receive a distribution of probabilities.

\section{Analytical Setup}
\subsection{Iterative Proportional Fitting Method}
The Iterative Proportional Fitting (IPF) method allows us to generate a synthetic micro-population for each CBG by aligning sample microdata (PUMS) with known marginal distributions (ACS). Our synthetic individuals are characterized by a set of five binary features known to influence travel:
\begin{itemize}
    \item \textit{household\_car\_share}: Whether household owns at least one car.
    \item \textit{individual\_senior}: Whether the person is 65 years or older.
    \item \textit{household\_income\_high}: Whether household income is $\geq \$100K$.
    \item \textit{individual\_employed}: Whether the person is currently employed.
    \item \textit{individual\_college}: Whether the person has a college degree or higher.
\end{itemize}
Each individual is thus uniquely positioned within the socio-demographic and spatial landscape of Seattle, enabling heterogeneous travel behavior simulation.

With original data matrix $M$ and marginal sum in $n$ dimensions $(u_1,u_2,...,u_n)$, the Iterative Proportional Fitting (IPF) algorithm could be performed to generate synthesized data $N$.

\begin{algorithm}[h!]
	\caption{IPF(M, u, $\epsilon$)}
  \hspace*{\algorithmicindent} \textbf{Input}  M: original matrix; $(u_1,u_2,...,u_D)$: marginal sum in each dimension\\ $\epsilon$: error tolerance\\
 \hspace*{\algorithmicindent} \textbf{Output} N: generated matrix
	\begin{algorithmic}[1]
            \State N = empty array with same shape as M
            \State $\delta$ = Euclidean distance between N and M 
		\While {$\delta$ $>$ $\epsilon$}
            \For {d in number of dimension of M}
		\State d\_sums = sum of each unit of M in dimension d
            \For {each element in N (represented as N[i,j,...,k,...], where k is the ordinal in dimension d)}
            \State N[i,j,...,k,...] = M[i,j,...,k,...] * $u_d$[k] / d\_sums[k]
            \EndFor
            \EndFor
		\EndWhile
	\end{algorithmic} 
\end{algorithm}

After scaling, the final result is a matrix containing decimal numbers, which must be integerized since the population should be represented as integers. The integerization process is crucial to maintain the discrete nature of the population count, and it is typically achieved using appropriate rounding or truncation methods. By integerizing the values, the results will be consistent with the expected whole-number population representation.

\subsection{Four-Step Modeling Framework}

\subsubsection{Trip Generation}
To predict travel volume based on demographic factors, several machine learning models are employed, including Random Forest, Linear Regression, Gradient Boosting, and Deep Learning. These models utilize one-hot encoding to represent ten different household and individual-level features from the PSRC dataset.
The machine learning models used in this study can be mathematically represented as follows:

\begin{itemize}
    \item \textbf{Linear Regression:}
    \[
    Y = \beta_0 + \sum_{i=1}^{n} \beta_i X_i + \epsilon
    \]
    where \( \beta_0 \) is the intercept, \( \beta_i \) are the coefficients for each feature \( X_i \), and \( \epsilon \) is the error term.

    \item \textbf{Random Forest:}
    \[
    \hat{Y} = \frac{1}{T} \sum_{t=1}^{T} f_t(\mathbf{X})
    \]
    where \( T \) is the number of trees, and \( f_t \) represents the prediction of the \( t \)-th decision tree.

    \item \textbf{Gradient Boosting:}
    \[
    \hat{Y} = \sum_{m=1}^{M} \gamma_m h_m(\mathbf{X})
    \]
    where \( M \) is the number of boosting stages, \( \gamma_m \) is the learning rate, and \( h_m \) represents the weak learner at stage \( m \).

    \item \textbf{Deep Learning:}
    \[
    \hat{Y} = f(\mathbf{W}, \mathbf{X})
    \]
    where \( f \) is the deep neural network function with weights \( \mathbf{W} \) and input features \( \mathbf{X} \).
\end{itemize}

To assess feature importance, we use models like Random Forest and Gradient Boosting. The importance of a feature in a Random Forest model, for example, can be quantified by the mean decrease in impurity (Gini importance) or permutation importance. Gradient Boosting models often use similar metrics to determine the influence of each feature on the prediction outcome.

The project also employs SHAP (SHapley Additive exPlanations) values to understand the contribution of each factor towards predicting travel volume. SHAP values provide a unified measure of feature importance by distributing the prediction difference fairly among the features, based on Shapley values from cooperative game theory.

Mathematically, SHAP values are calculated as:

\[
\phi_i = \sum_{S \subseteq \{1, \ldots, n\} \setminus \{i\}} \frac{|S|!(n - |S| - 1)!}{n!} \left[ f(S \cup \{i\}) - f(S) \right]
\]

where \( \phi_i \) is the SHAP value for feature \( i \), \( S \) is a subset of all features excluding \( i \), \( f(S) \) is the model prediction for the subset \( S \), and \( n \) is the total number of features. This ensures that each feature's contribution is fairly attributed, considering all possible combinations of features.

\begin{figure}[htbp]
    \centering
    \includegraphics[width=0.45\textwidth]{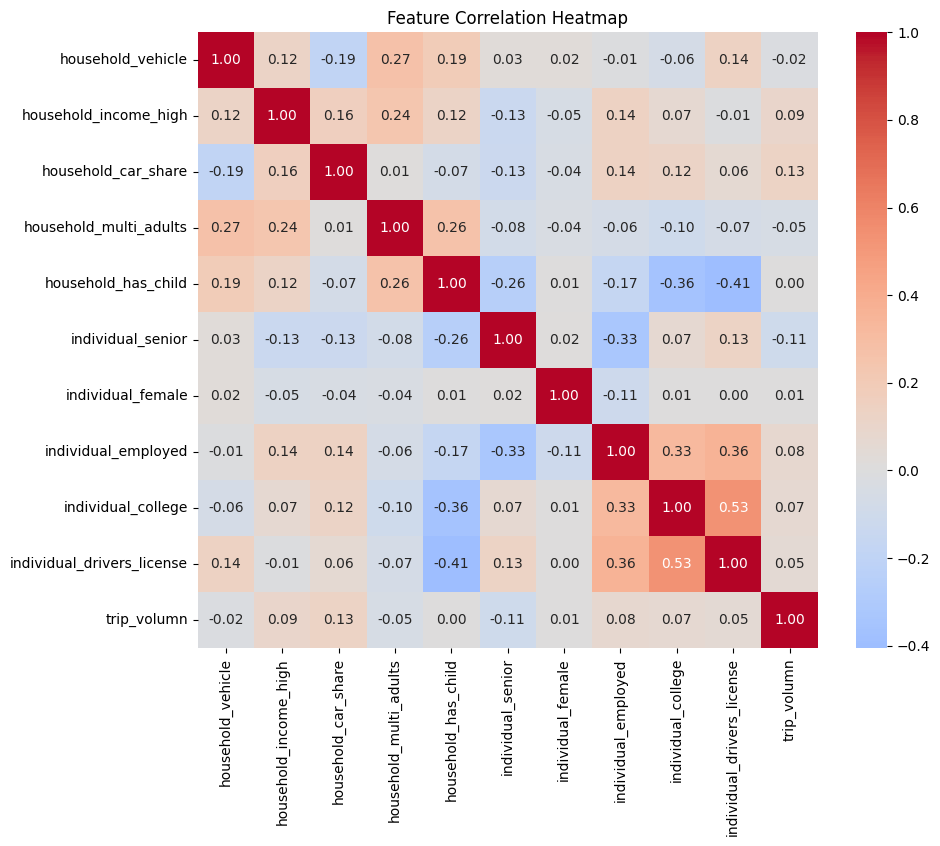}  %
    \includegraphics[width=0.45\textwidth]{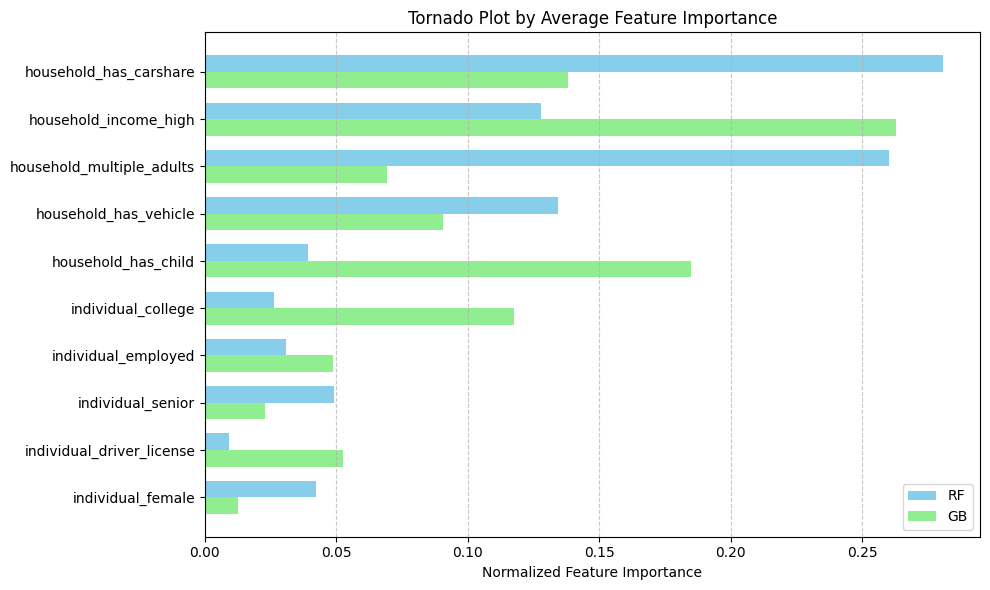}
    \caption{Feature cross correlation and importance analysis}
    \label{fig:feature}
\end{figure}

In Figure \ref{fig:feature}, we observe the importance of different features across various machine learning models. SHAP values illustrate how each factor's significance varies depending on the specific prediction scenario. This detailed analysis of feature importance is crucial for understanding the underlying drivers of travel volume and aids in making informed decisions regarding transportation policies and infrastructure planning.

\subsubsection{Trip Distribution}

Trip distribution estimates the number of trips between origin-destination (OD) pairs, forming the essential bridge between trip generation and route assignment. One widely adopted analytical approach for trip distribution is the gravity model, which draws conceptual inspiration from Newtonian physics while being grounded in statistical mechanics through entropy maximization. From a transportation systems perspective, the entropy-maximizing formulation offers a principled framework that balances trip productions, trip attractions, and travel impedance in a probabilistically optimal manner.

Mathematically, the objective is to determine the trip matrix $\{T_{ij}\}$ that maximizes the overall entropy of the system, reflecting the number of possible microstates (i.e., trip configurations) consistent with observed constraints. The entropy function is defined as:
\[
S = - \sum_{i,j} T_{ij} \ln T_{ij},
\]
subject to the marginal constraints:
\[
\sum_j T_{ij} = O_i, \quad \sum_i T_{ij} = D_j,
\]
and optionally a cost constraint or impedance function that penalizes long-distance trips. By applying the method of Lagrange multipliers, the resulting optimal solution yields the familiar doubly-constrained gravity model form:
\[
T_{ij} = A_i B_j O_i D_j f(c_{ij}),
\]
where $f(c_{ij})$ is a deterrence function that decreases with travel cost $c_{ij}$, and $A_i$, $B_j$ are balancing factors ensuring that the marginal constraints are satisfied. Common choices for the deterrence function include exponential decay $f(c_{ij}) = e^{-\beta c_{ij}}$ and power functions $f(c_{ij}) = c_{ij}^{-\gamma}$, where $\beta$ and $\gamma$ are impedance parameters estimated from data.

This entropy-based interpretation not only provides a statistical underpinning to the gravity model but also facilitates the incorporation of additional constraints or policy levers, such as congestion pricing, accessibility thresholds, or socioeconomic factors. Its widespread adoption in metropolitan planning organizations (MPOs) and regional modeling frameworks is attributed to its analytical tractability, interpretability, and empirical robustness. Moreover, the entropy maximization perspective aligns naturally with emerging data sources (e.g., mobile phone records, GPS traces) by allowing model calibration using maximum likelihood or Bayesian updating techniques. As such, it remains a cornerstone methodology for estimating plausible OD matrices under constrained conditions, particularly in the absence of fully observed trip-level data.

\subsubsection{Route Assignment}

Routes for each origin-destination (OD) pair are determined using Dijkstra’s shortest path algorithm applied to a multimodal transportation graph composed of road links, walking paths, and GTFS-derived transit edges. This unified network structure allows seamless modeling of complex travel behaviors, including walk-to-transit and intermodal transfers. Edge weights are defined in terms of generalized impedance, which may include travel time, transfer penalties, wait times, and real-time congestion adjustments when available. This enables flexible simulation of realistic multimodal flows and link-level load estimations.

Dijkstra’s algorithm is a fundamental and well-established method for solving the single-source shortest path problem in graphs with non-negative edge weights. It incrementally builds the shortest path tree by repeatedly selecting the node with the smallest known tentative distance from the source, updating the distances of its neighbors, and finalizing nodes whose minimum paths are guaranteed. Its correctness and determinism make it a reliable foundation for routing under general cost metrics. The algorithm proceeds as follows:

\begin{algorithm}
    \caption{Dijkstra(Graph, source)}
    \hspace*{\algorithmicindent} \textbf{Input} Graph: a weighted graph with nodes and edges; source: starting node\\
    \hspace*{\algorithmicindent} \textbf{Output} dist[]: shortest distance from source to all nodes; prev[]: previous node in optimal path
    \begin{algorithmic}[1]
        \State Initialize dist[source] = 0
        \For {each node $v$ in Graph}
            \If {$v \ne$ source}
                \State dist[$v$] = $\infty$
            \EndIf
            \State prev[$v$] = undefined
        \EndFor
        \State Q = all nodes in Graph (min-priority queue based on dist[])
        \While {Q is not empty}
            \State $u$ = node in Q with smallest dist[$u$]
            \State remove $u$ from Q
            \For {each neighbor $v$ of $u$}
                \State alt = dist[$u$] + weight($u$, $v$)
                \If {alt < dist[$v$]}
                    \State dist[$v$] = alt
                    \State prev[$v$] = $u$
                \EndIf
            \EndFor
        \EndWhile
    \end{algorithmic}
\end{algorithm}

The algorithm yields the shortest paths from the source to all reachable nodes, which can then be traced using the \texttt{prev[]} mapping. Within a multimodal urban network, the edge weights \texttt{weight(u, v)} may include static or dynamic impedance terms such as congested travel times, walk speeds, transit headways, and transfer costs. This generality makes the algorithm particularly well-suited for the task. For small to medium-sized networks, such as district-scale simulations, Dijkstra’s algorithm is computationally efficient and scalable, especially when paired with appropriate data structures like binary heaps or adjacency lists.


\subsubsection{Mode Choice}
The Naive Bayesian approach provides a robust statistical framework for understanding the probabilities associated with different transportation modes based on demographic data and other relevant factors.

The Naive Bayesian Classifier simplifies the computation by assuming that the demographic features are conditionally independent given the transportation mode. This assumption, although often violated in practice, allows for efficient and effective computation. The fundamental equation of the Naive Bayesian classifier in our context is expressed as:

\[
P(T|D_1, D_2, \ldots, D_n) \propto P(T) \cdot \prod_{i=1}^{n} P(D_i|T)
\]

where \( T \) represents the transportation mode (e.g., transit, drive, walk), and \( D_i \) represents individual demographic attributes such as race, sex, income, etc.

The steps for implementing the Naive Bayesian Classifier in our transportation mode prediction are as follows:

\begin{itemize}

\item Compute Prior \( P(T) \):
   We start by calculating the prior probabilities for each transportation mode from the training dataset:

\[
P(T) = \frac{\text{Number of trips using mode } T}{\text{Total number of trips}}
\]

\item Estimate Likelihood \( P(D_i|T) \):
   For each demographic attribute \( D_i \), we estimate the likelihood of observing \( D_i \) given transportation mode \( T \):

\[
P(D_i|T) = \frac{\text{Number of individuals with } D_i \text{ using mode } T}{\text{Total number of individuals using mode } T}
\]

\item Calculate Posterior \( P(T|D) \):
   Using the computed prior and likelihood, we calculate the posterior probability for each transportation mode given the demographic data:

\[
P(T|D_1, D_2, \ldots, D_n) \propto P(T) \cdot \prod_{i=1}^{n} P(D_i|T)
\]

\item Mode Choice Prediction:
   The transportation mode with the highest posterior probability is selected as the predicted mode for the given demographic profile:

\[
\hat{T} = \arg \max_T \, P(T|D_1, D_2, \ldots, D_n)
\]

\end{itemize}

We validate our Naive Bayesian Classifier by comparing the predicted mode choices with the actual mode choices from ground truth data. This comparison provides insights into the accuracy and reliability of the model. The Bayesian model's predictions enable better decision-making regarding transportation policies and infrastructure. By understanding the probabilistic relationships between demographics and transportation modes, policymakers can design targeted interventions to improve the efficiency and equity of transportation systems. The model's capability to integrate diverse data sources and produce granular predictions at the census block group level further enhances its utility for urban planning and policy formulation.

\section{Evaluation and Results}
\subsection*{Real-time Visualization and User Interface}

\begin{figure}[htbp]
    \centering
    \includegraphics[width=0.5\textwidth]{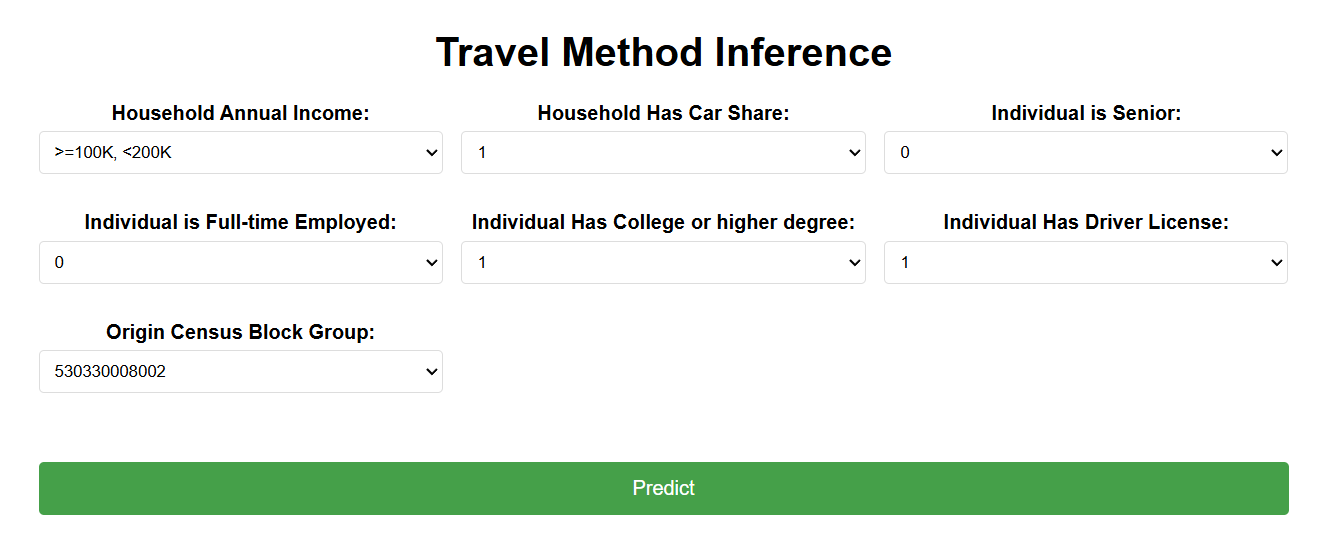}  
    \caption{Input Boxes in Our Website}
    \label{fig:input}
\end{figure}

\begin{figure}[htbp]
    \centering
    \includegraphics[width=0.4\textwidth]{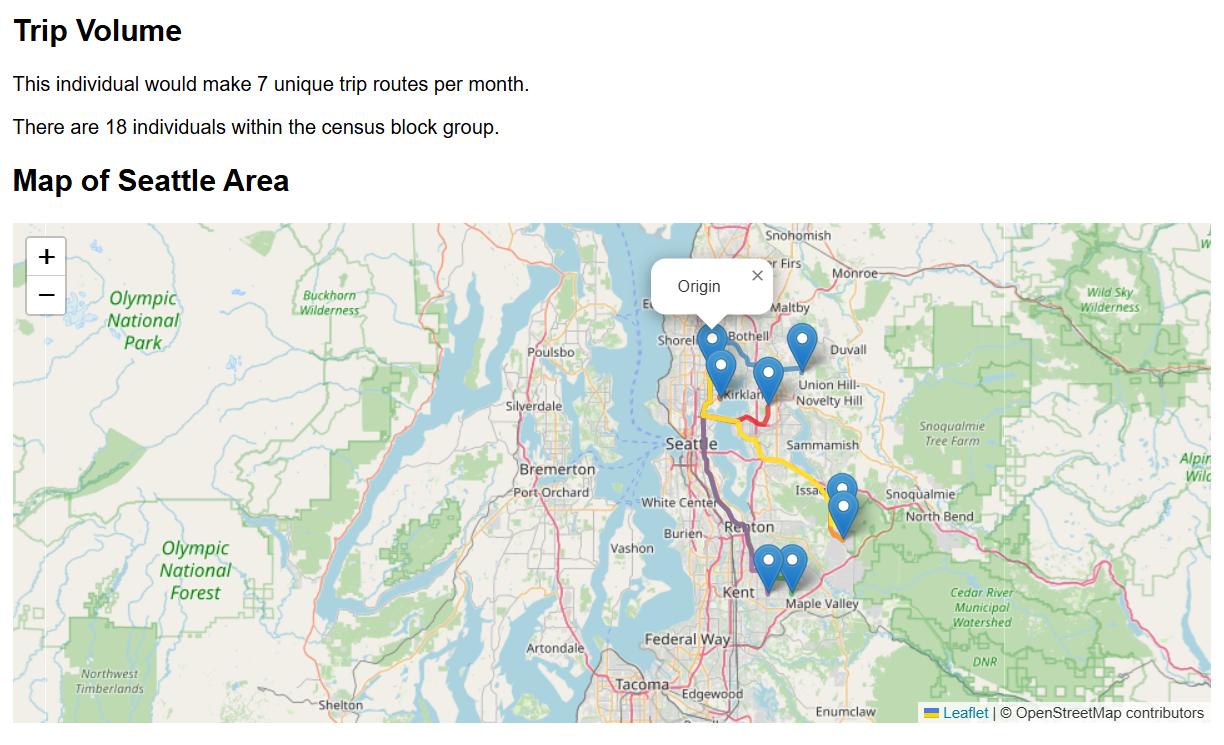}
    \includegraphics[width=0.4\textwidth]{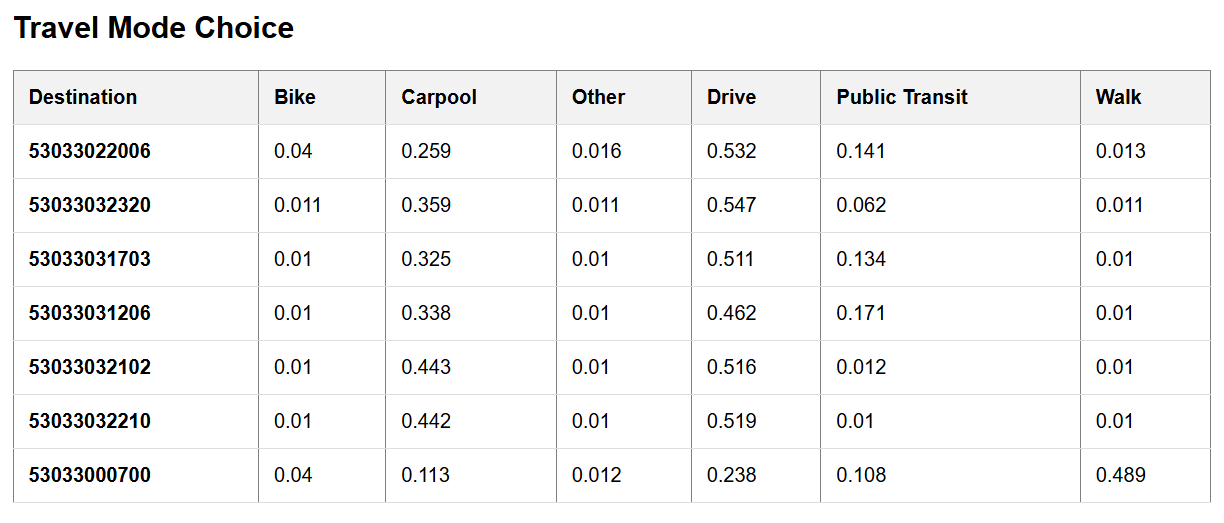}
    \caption{Result Outputs Once Clicked 'Predict'}
    \label{fig:output}
\end{figure}

This project presents an analytical framework, manifested as a user-friendly web interface, to explore and predict individual travel behaviors and transportation choices. By allowing planners and policymakers to input a set of demographic and household variables, such as income, car ownership, and personal characteristics like age and education level, the system simulates and reveals latent mobility patterns. At its core, the application employs a series of predictive models trained on extensive regional transportation data to sequentially infer a person's monthly trip volume, identify probable destinations, and, most critically, calculate the likelihood of different travel modes—such as driving, public transit, or walking—for each journey. The visual output, including a map-based representation of potential routes and a clear breakdown of modal probabilities, provides a powerful tool for understanding how socioeconomic factors and spatial relationships influence human movement. This approach shifts the focus from simple data reporting to an interactive predictive analysis, empowering stakeholders to gain data-driven insights into urban mobility dynamics and to make more informed decisions regarding transportation policy, infrastructure investment, and sustainable urban planning.

\subsection*{Robustness and Validation}

The validation process is performed at the "census block group level". ACS/PUMS data, particularly when linked to census block groups, provides rich demographic and socioeconomic characteristics. This means that validation is not just about assessing overall regional accuracy but critically about evaluating accuracy within specific neighborhoods and for distinct demographic segments. This allows for a detailed spatial and social assessment of model performance. This granular validation capability allows for the identification of specific census block groups or demographic groups where the model might perform less accurately, thereby highlighting areas that require further data collection, model refinement, or targeted calibration. More importantly, it enables the assessment of whether proposed transportation solutions (informed by the model's predictions) are equitable in their impact across different communities, directly aligning with the project's implicit goal of aiding effective transportation planning and policy-making for all residents of Seattle.

\begin{figure}[htbp]
    \centering
    \includegraphics[width=0.4\textwidth]{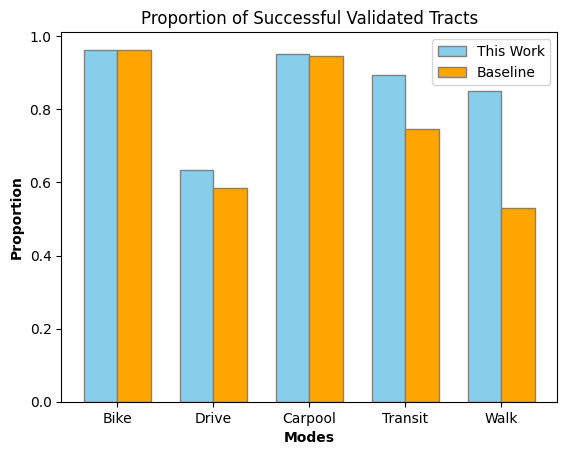}  
    \caption{Proportion of Correctly Predicted Census Block Groups}
    \label{fig:val}
\end{figure}

To validate our results, we combine the population of individuals in each census block group and estimate the transportation volume, subsequently comparing these estimates with those generated by the baseline approach, which is based on simple population proportion.

The naive baseline approach distributes the proportion of individuals using each transportation method uniformly across all census blocks. This method assumes homogeneity within every census block group, disregarding variations in demographic distribution, origin-destination patterns, and other critical factors that influence transportation choices. By ignoring these variations, the baseline method provides a rudimentary and often inaccurate representation of actual transportation volumes.

In contrast, our approach accounts for the heterogeneity and spatial nuances of the population, significantly enhancing the accuracy and granularity of transportation volume estimates. This is evident in the model's performance, which outstrips the baseline methods, particularly in detecting low accuracy in transit, drive, and walk modes. The results, illustrated in Figure \ref{fig:val}, demonstrate the model's superior capability in capturing the intricacies of travel behavior within diverse urban environments.

One of the key strengths of our model is its ability to provide data at a finer granularity, specifically targeting census block groups rather than the broader census tracts used in the American Community Survey (ACS). Most existing data sources only offer information at the census block group level, limiting the precision and applicability of the data for localized policy and planning purposes. Our approach, however, breaks down the data into more detailed census block groups, allowing for a more nuanced understanding of transportation patterns and better-informed decision-making.

The significance of this project's advancements lies not only in its improved accuracy but also in its potential to offer insightful policy implications. By delivering more granular and precise data, the model enables policymakers to identify and address specific transportation challenges within smaller geographic areas. This level of detail is particularly valuable for urban planners and local authorities who need to design targeted interventions to enhance transportation efficiency and equity.

\section{Policy Insights and Conclusions}

The proposed framework offers a flexible and actionable toolset for local governments, Metropolitan Planning Organizations (MPOs), and transit agencies aiming to modernize their transportation planning practices. By enhancing the classical four-step model with disaggregated agent-based inputs and a multimodal routing engine, the methodology supports a wide array of traffic engineering and policy use cases. In terms of infrastructure planning, it enables identification of multimodal bottlenecks and high-stress corridors, providing evidence for prioritizing investments such as dedicated bus lanes, pedestrian safety upgrades, or signal timing improvements. The framework also facilitates equity-aware analysis by allowing planners to audit modal access and travel burdens across socio-demographic strata, including income groups, elderly populations, and zero-vehicle households.

A key contribution of this work lies in its ability to detect and quantify travel disadvantage at the population level. By modeling individuals and households explicitly, the system surfaces critical spatial and modal mismatches that are often obscured in aggregate models. For example, it highlights elderly residents without access to private vehicles living in transit-sparse neighborhoods, low-income workers facing long and multimodal commutes, or communities where affordable housing is disconnected from major employment centers. These findings can inform targeted interventions such as the deployment of microtransit services, expansion of transit coverage zones, or development incentives aligned with mobility goals.

In a broader context, this work demonstrates how traditional transportation modeling can be revitalized through the integration of multi-source datasets, statistical estimation techniques, and network-theoretic algorithms. The grounding of the four-step process in synthetic microdata allows planners to reconcile the rigor of system-level demand modeling with the nuance of individual mobility needs. As transportation systems continue to evolve toward multimodality, equity, and sustainability, such granular and adaptive modeling approaches will be increasingly vital. This framework provides a pathway to bridge long-standing gaps between regional modeling tools and the realities of everyday travel, ultimately supporting more equitable, data-informed, and human-centric urban mobility systems.

\end{document}